\def\year{2019}\relax
\documentclass[letterpaper]{article} 
\usepackage{aaai19}  
\usepackage{times}  
\usepackage{helvet}  
\usepackage{courier}  
\usepackage{url}  
\usepackage{graphicx}  

\usepackage[usenames, dvipsnames]{xcolor}
\usepackage{multirow}
\usepackage{algorithm}
\usepackage{algorithmic}
\usepackage{amsfonts}
\usepackage{mathtools,amssymb}
\usepackage{subcaption}
\newcommand{\eqn}[1]{Equation~#1}
\newcommand{\fig}[1]{Fig.~#1}
\newcommand{\tab}[1]{Table~#1}

\usepackage{expl3}
\ExplSyntaxOn
\newcommand\latinabbrev[1]{
	\peek_meaning:NTF . {
		#1\@}%
	{ \peek_catcode:NTF a {
			#1.\@ }%
		{#1.\@}}}
\ExplSyntaxOff
\def\eg{\latinabbrev{e.g}}
\def\etal{\latinabbrev{et al}}
\def\etc{\latinabbrev{etc}}
\def\ie{\latinabbrev{i.e}}

\begin{document}
%
\title{MVPNet: Multi-View Point Regression Networks for 3D Object Reconstruction from A Single Image}

\author{
	Jinglu Wang\footnotemark[2] \quad Bo Sun\footnotemark[3]\thanks{The work was done when Bo Sun was an intern at MSR.} \quad Yan Lu\footnotemark[2]\\
	\footnotemark[2]Microsoft Research \quad \footnotemark[3]Peking University\\
	\{jinglwa,~v-bosu,~yanlu\}@microsoft.com
}

\maketitle
\begin{abstract}
In this paper, we address the problem of reconstructing an object's surface from a single image using generative networks. 
First, we represent a 3D surface with an aggregation of dense point clouds from multiple views. Each point cloud is embedded in a regular 2D grid aligned on an image plane of a viewpoint, making the point cloud convolution-favored and ordered so as to fit into deep network architectures. The point clouds can be easily triangulated by exploiting connectivities of the 2D grids to form mesh-based surfaces. Second, we propose an encoder-decoder network that generates such kind of multiple view-dependent point clouds from a single image by regressing their 3D coordinates and visibilities.
We also introduce a novel geometric loss that is able to interpret discrepancy over 3D surfaces as opposed to 2D projective planes, resorting to the surface discretization on the constructed meshes. 
We demonstrate that the multi-view point regression network outperforms state-of-the-art methods with a significant improvement on challenging datasets.
\end{abstract}

\section{Introduction}
\begin{figure}[t]
	\centering
	\includegraphics[width=0.94\linewidth,trim={0 0cm 0cm 0cm}, clip ]{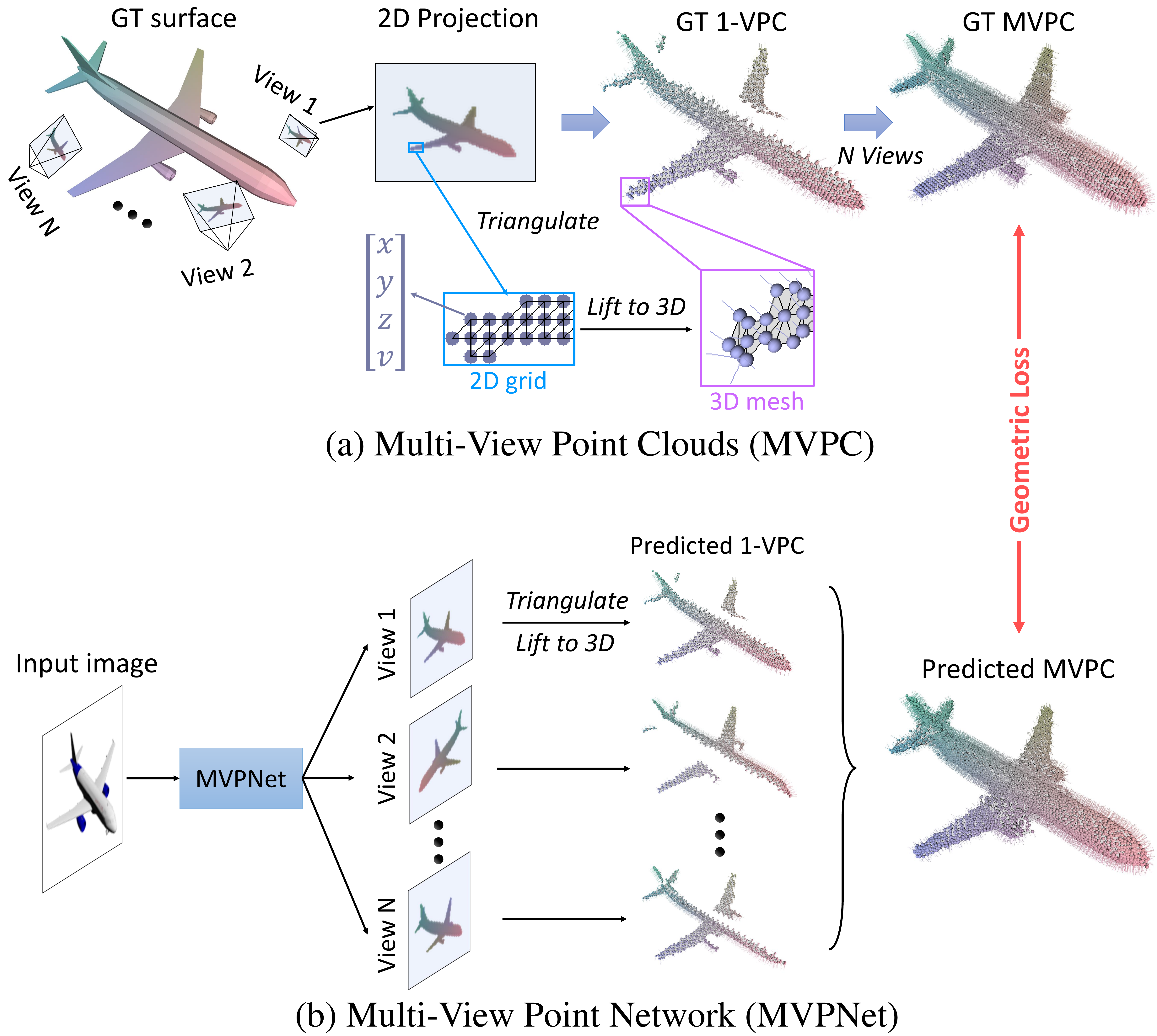}
	\caption{(a) A surface is represented by MVPC. Each pixel in a 1-VPC stores the backprojected surface point $(x,y,z)$ from this pixel and its visibility $v$. The stored 3D points are triangulated according to the 2D grid on the image plane and their normals are shown to indicate surface orientation. 
(b) Given an RGB image, the MVPNet generates a set of 1-VPCs and their union forms the predicted MVPC. The geometric loss measures discrepancy between predicted and groundtruth MVPC.
}
	\label{fig:overview_aaai}
\end{figure}
3D object reconstruction from a single RGB image is an inherently ill-posed problem as many configurations of shape, texture, lighting, and camera can give rise to the same observed image.  
Recently, the advanced deep learning models allow for the rethinking of this task as generating realistic samples from underlying distributions. 
Regular representations are favored by deep convolutional neural networks for dense data sampling, weight sharing, \etc. Although meshes are the predominant representations for 3D geometries, their irregular structures are not easy for encoding and decoding.
Most extant deep nets \cite{choy20163dr2n2,tulsiani2017drc,wu20163dgan,zhu2017rethinking,girdhar2016tlembedding} employ 3D volumetric grids. However, they suffer from high computational complexity for dense sampling. 
A few recent methods \cite{fan2017point,achlioptas2017representation} advocate the unordered point cloud representation. The unordered property requires additional computation to establish a one-to-one mapping for point pairs. It often yields sparse results because of costly mapping algorithms. 

In order to depict dense and detailed surfaces, we introduce an efficient and expressive view-based representation inspired by recent studies on multi-view projections~\cite{kalogerakis2017projective,soltani2017synthesizing,shin2018pixels}.
In particular, we propose to represent a surface by dense point clouds visible from multiple viewpoints. The arrangement of viewpoints are configured to cover most of the surface.
The multi-view point clouds (MVPC) are illustrated in \fig{\ref{fig:overview_aaai}} (a). Each point cloud is stored in a 2D grid embedded in a viewpoint's image plane.
A 1-view point cloud (1-VPC) looks like a depth map, but each pixel stores the 3D coordinates and visibility information rather than the depth of the backprojected surface point from this pixel. The backprojection transformation offers a one-to-one mapping of point sets in 1-VPCs with equal camera parameters.
Meanwhile, local connectivities of the 3D points are introduced from the 2D grids, which facilitate to form a triangular mesh based on such backprojected points. 

Accordingly, the surface reconstruction problem is formulated as the regression of values stored in MVPC.
We employ an encoder-decoder network as a conditional sampler to generate the underlying MVPC, as shown in \fig{\ref{fig:overview_aaai}} (b). The encoder extracts image features and combines them with different viewpoints' features respectively. The decoder consists of multiple weight-shared branches, each of which generates a view-dependent point cloud. The union of all 1-VPCs forms the final MVPC. We propose a novel \emph{geometric loss} that measures discrepancies over real 3D surfaces as opposed to 2D planes. 
Unlike previous view-based methods processing features in 2D projective spaces (\ie, image planes) and neglecting the information loss through dimension reduction from 3D to 2D, the proposed MVPC allow us to discretize integrals of surface variations over the constructed triangular mesh.
The geometric loss integrating volume variations, prediction confidences and multi-view consistencies contributes to high reconstruction performance.

\vspace{-0.2cm}
\section{Related Work}

\subsubsection{Mesh-based methods.} 
Mesh representation has been extensively used to improve and manipulate surface interfaces. In particular, surface reconstruction is usually posed to deform an initial mesh to minimize a variational energy functional in the spirit of data fidelity. \cite{delaunoy2011gradient,pons2007multi,wang2016image,liu2015higher} are the pioneers of reconstructing mesh-based surface from multi-views. These deformable mesh methods calculate the integral over the whole surface and thus capture complete properties on the surface.
However, irregular connectivities of mesh representation make it difficult to leverage the advance of convolutional architectures. Recent methods~\cite{pontes2017image2mesh} use linear combinations of a dictionary of CAD models and learn the parameters of the combination to represent the models, which are limited to the capacity of the constructed dictionary.
We are inspired by variational methods that have geometric interpretations for optimization formulation. Important geometric clues are integrated into the loss function, which contributes a superior performance significantly.          

\subsubsection{Voxel-based methods.}
When learning methods dominate the recognition tasks, volumetric representation~\cite{girdhar2016tlembedding,wu20163dgan,choy20163dr2n2,wu2017marrnet,tulsiani2017drc} is more favored because of its regular grid-like structure that suites convolutional operations.
Tulsiani and Zhou~\cite{tulsiani2017drc} formulate a differentiable ray consistency term to enforce view consistency on the voxels with the supervision of multi-view observation.
3D-R2N2~\cite{choy20163dr2n2} learns to aggregate voxel occupancy from sequential input images and can obtain robust results. Voxel-based methods are limited by the cubic growth rate of both memory and computation time, leading to low-resolution of grids. 

\subsubsection{View-based methods.}
As the drawbacks of voxel-based CNNs are obvious, some methods adopt view-based representations. 
They project surfaces on image planes with regular 2D grids that allows planar convolution.
A few methods~\cite{park2017transformation,zhou2016view,chen2018stereoscopic,chen2017stylebank} achieve impressive results in synthesizing novel views from a single view.
Tatarchenko \etal \cite{tatarchenko2016mv3d} utilize CNNs to infer images and depth maps of arbitrary views given an RGB image, and then fuse the depth maps to yield a 3D surface. 
Soltani \etal \cite{soltani2017synthesizing} synthesize multi-view depth maps from a single or multiple depth maps. Since depth maps inherently contain geometric information, our task, which takes a single RGB image as an input is much more challenging.
Lin \etal \cite{lin2017learning} also generate points of multiple views with a generative network.
These methods all focus on predicting the intermediate information in 2D projective planar spaces yet ignore real 3D spatial correlation and multi-view consistency.
Our method incorporates the spatial correlation of surface points and further enforces multi-view consistency to achieve more accurate and robust reconstructions.

\subsubsection{Point-based methods.}
Some methods generate an unordered point cloud from an image by deep learning.
Su \etal \cite{fan2017point} are the first to study the problem. The unordered property of a point cloud enjoys high flexibility~\cite{qi2017pointnet}, but it increases computational complexity due to lack of correspondences. This makes such methods not scalable, resulting in sparse points.

\vspace{-0.1cm}
\section{Approach}
In this section, we first formally introduce the MVPC representation for depicting 3D surfaces efficiently and expressively. Then, we detail the MVPNet architecture and the geometric loss for generating the underlying MVPC conditioned on an input image.
\vspace{-0.1cm}
\subsection{MVPC Representation}
\label{sec:representation}
\begin{figure}[t]
	\centering
	\includegraphics[width=0.96\linewidth,trim={0 0cm 0cm 0cm},clip]{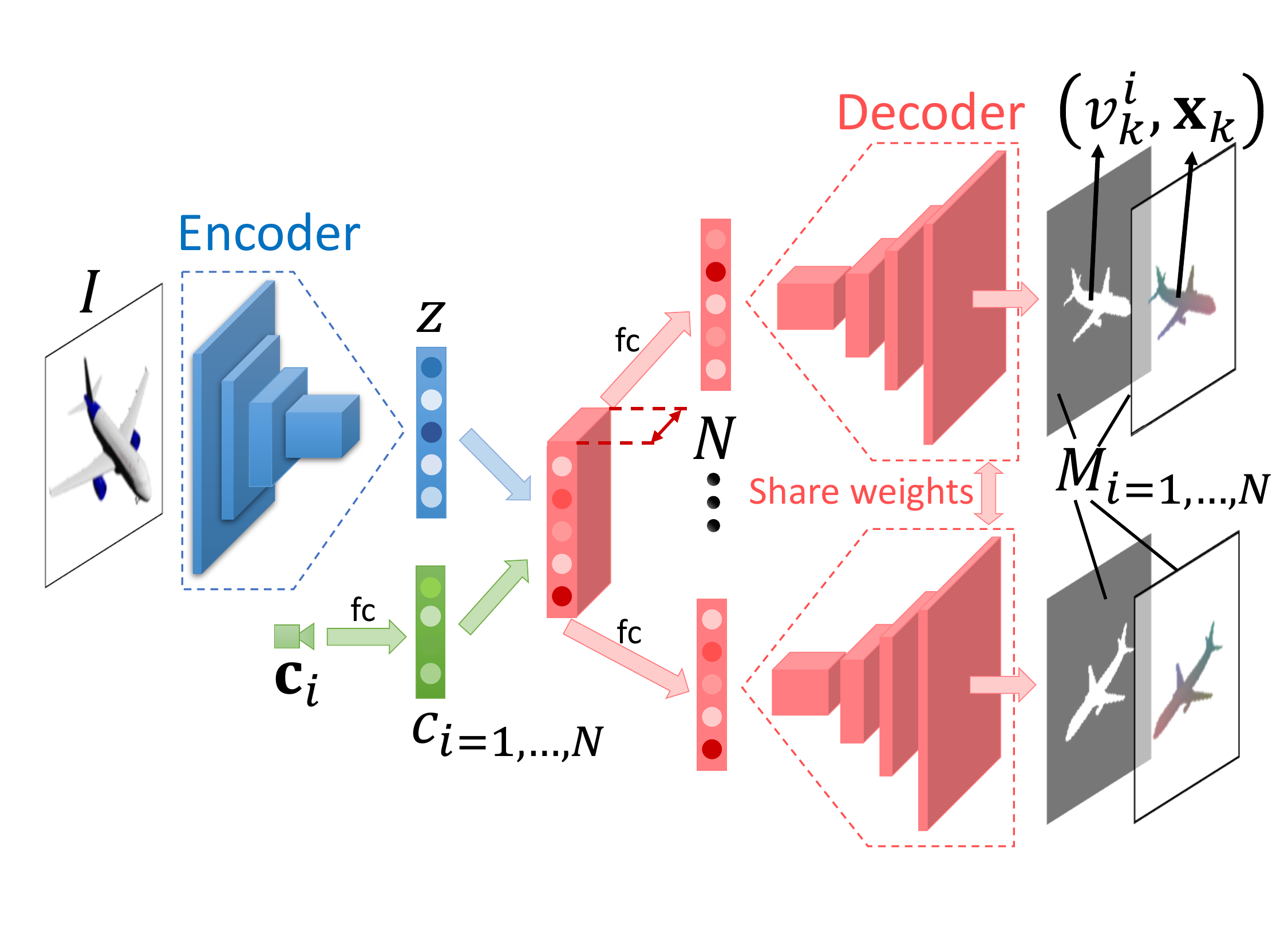}
	\caption{\textbf{MVPNet architecture.} Given an input image $I$, MVPNet consisting of an encoder and a decoder regresses the $N$ 1-VPCs $\{M_i\}$ for $\{\mathbf{c}_i\}$, $i=1,..,N$ respectively. $N$ concatenated features $(z, c_i)$ are fed into $N$ branches of the decoder, of which the branches share weights.
	}
	\label{fig:net}
\end{figure}

An object's surface $\mathcal{S}$ is considered as an aggregation of partial surfaces $\bigcup_{i=1}^N S_i$ visible from a set of predefined viewpoints $\{\mathbf{c}_i|i=1,...,N\}$. Each partial surfaces $S_i$ is discretized and parameterized by the aligned 2D grid on the image plane of $\mathbf{c}_i$, as shown in \fig{\ref{fig:overview_aaai}} (a).
Each pixel $x_k$ on the grid stores the 3D point $\mathbf{x}_k = (x_k,y_k,z_k)$ backprojected from $x_k$ onto $\mathcal{S}$ and the visibility $v_k^i$ of $\mathbf{x}_k$ from $\mathbf{c}_i$. $v_k^i$ is set to 1 if $\mathbf{x}_k$ is visible from $\mathbf{c}_i$, otherwise $0$.
The visible 3D points are triangulated by connecting them with the 2D grid's horizontal, vertical, and one of the diagonal edges to form a mesh-based surface.
Such multiple view-dependent parameterized surfaces are named multi-view point clouds, MVPC in short, $\mathcal{M}=\bigcup_{i=0}^N M_i$, where $M_i$ denotes a 1-view point cloud, 1-VPC in short.
Let $\mathcal{X} = \{\mathbf{x}_k \}$ denote all the 3D points in $\mathcal{M}$.

MVPC inherit the advantage of efficiency from general view-based representations. Unlike volumetric representation using costly 3D convolution, 2D convolution is performed on the 2D grids, which encourages higher resolutions for denser surface sampling.
Meanwhile, MVPC encode one-to-one mapping of predicted points $\mathcal{X}$ and groudtruth points $\widetilde{\mathcal{X}}$ explicitly. 
Induced by the same viewpoint $\mathbf{c}_i$, pixels with the same 2D coordinates $x_k$ are defined to store the same surface point $\mathbf{x}_k$. In other words, the groundtruth and predicted 1-VPC of the same viewpoint store the points in the same order. Compared to unordered point cloud representations that require additional computation to construct point-wise mapping, MVPC have superior performance in computation.

MVPC express not a simple combination of multi-view projections but a discrete approximation to a real 3D surface.
On the one hand, A triangular mesh is constructed for each 1-VPC, and thus we can formulate losses based on geometries on 3D surfaces rather than on 2D projections. 
Note that the edges inherited from 2D grids are not all real in 3D, \eg, edges connecting points on depth discontinuities are fake. We deal with the fake edges by penalizing them largely in the loss formulation. 
On the other hand, we carefully select relatively few yet evenly distributed viewpoints on a viewing sphere that can cover most of the targeting surface. Different numbers of viewpoints are discussed in the experiment section. We also consider multi-view consistency constraints in overlap regions to improve the expressiveness of MVPC.

\subsection{MVPNet Architecture}
\label{sec:network}
We exploit an encoder-decoder generative network architecture and incorporate camera parameters into the network to generate view-dependent point clouds. The network architecture is illustrated in \fig{\ref{fig:net}}.
The encoder learns to map an image $I$ to an embedding space to obtain a latent feature $z$. Each camera matrix $\mathbf{c}_i$ is first transformed to a higher-dimensional hidden representation $c_i$, serving as a view indicator, and then is concatenated with $z$ to get $(z,c_i)$. The decoder that converts $(z,c_i)$ to a 1-VPC $M_i$ indicated $\mathbf{c}_i$ learns the projective transformation and space completion. The decoder shares weights among $N$ branches. The output MVPC $\mathcal{M}=\bigcup_{i=1}^N M_i$ is of shape $N \times H \times W \times 4$, where $H$ and $W$ denote the height and width of a 1-VPC. The last channel corresponds to a 3D coordinate $\mathbf{x}_k=(x_k,y_k,z_k)$ and visibility $v_k^i$ of a point $\mathbf{x}_k$. 

The encoder is a composition of convolution and leaky ReLU layers. The camera parameters are encoded with fully connected layers. The decoder contains a sequence of transposed-convolution and leaky ReLU layers. The last layer is activated with the tanh functions, responsible for regressing 3D coordinates and visibilities of points. Implementation details are described in the experiment section. 


\vspace{-0.4cm}
\subsection{Geometric Loss}
\label{sec:loss}
While most point generation methods~\cite{fan2017point,lin2017learning,soltani2017synthesizing} adopt point-wise distance metrics, they disregard geometric characteristics of surfaces.
These networks attempt to predict ``mean'' shapes~\cite{fan2017point}, failing to preserve fine details.

We propose a geometric loss (GeoLoss) that is able to capture variances over 3D surfaces rather than over sparse point sets or 2D projective planes. We expand the GeoLoss to be differentiable for neuron networks and also to be robust against noise and incompletion.
The GeoLoss is made up of three components:
\begin{eqnarray}
\mathcal{L}_{Geo} = \mathcal{L}_{ptd} + 
\alpha \mathcal{L}_{vol} + \beta \mathcal{L}_{mv}
\end{eqnarray}
where $\mathcal{L}_{ptd}$ is the sum of distances between corresponding point pairs, $\mathcal{L}_{vol}$ denotes the quasi-volume term measuring discrepancy of local volumes, and $\mathcal{L}_{mv}$ is the multi-view consistency term. Coefficients $\alpha$ and $\beta$ are the weights balancing different losses.
\subsubsection{Point-wise distance term.}
\begin{figure*}[t]
	\centering
	\scriptsize
	\begin{tabular}{c@{\hspace{1cm}}c@{\hspace{0.5cm}}c}
		
	\includegraphics[height=0.2\linewidth, trim={0cm 1cm 0cm 0cm}, clip]{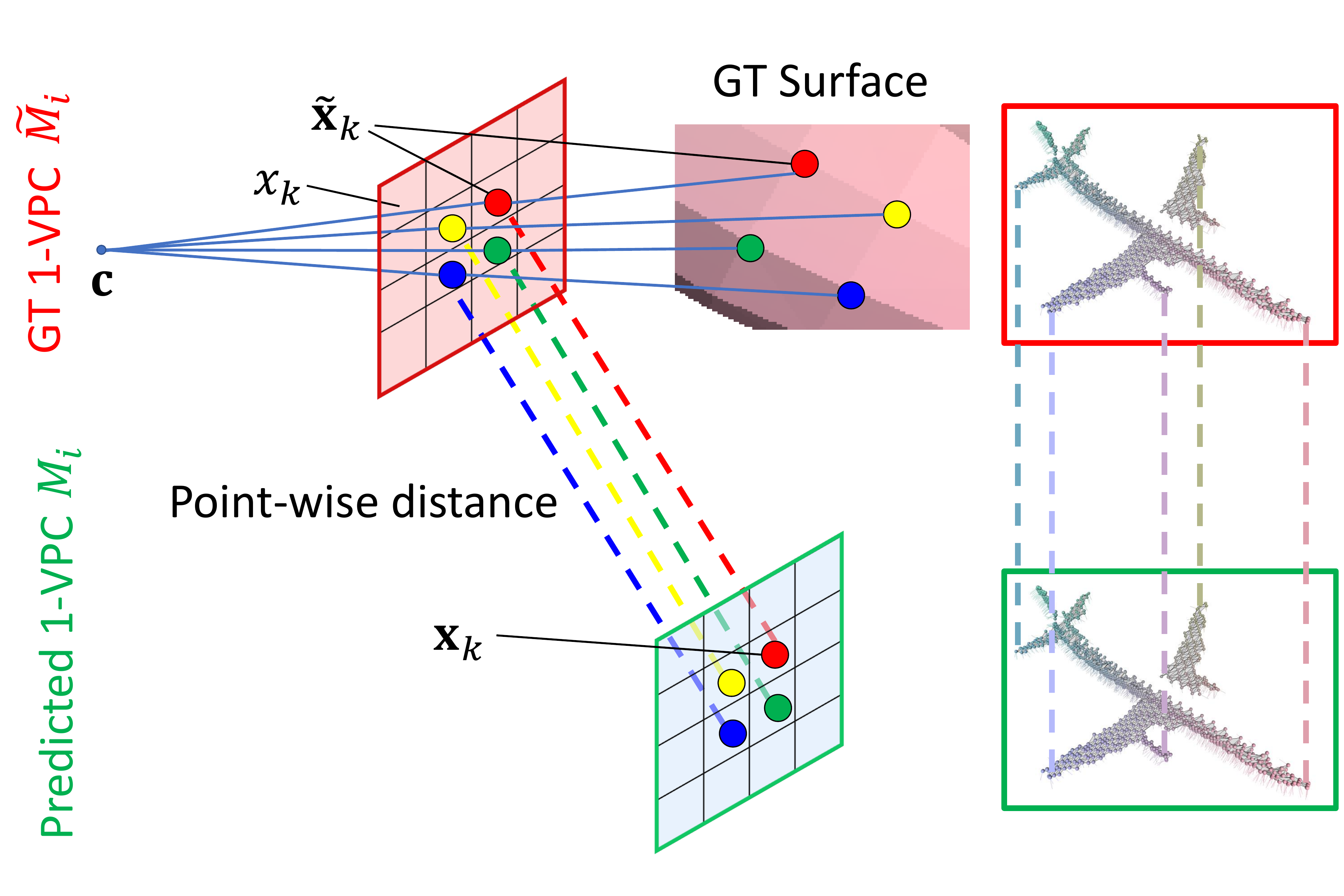} &
	\includegraphics[height=0.2\linewidth, trim={0cm 2cm 0cm 0cm}, clip]{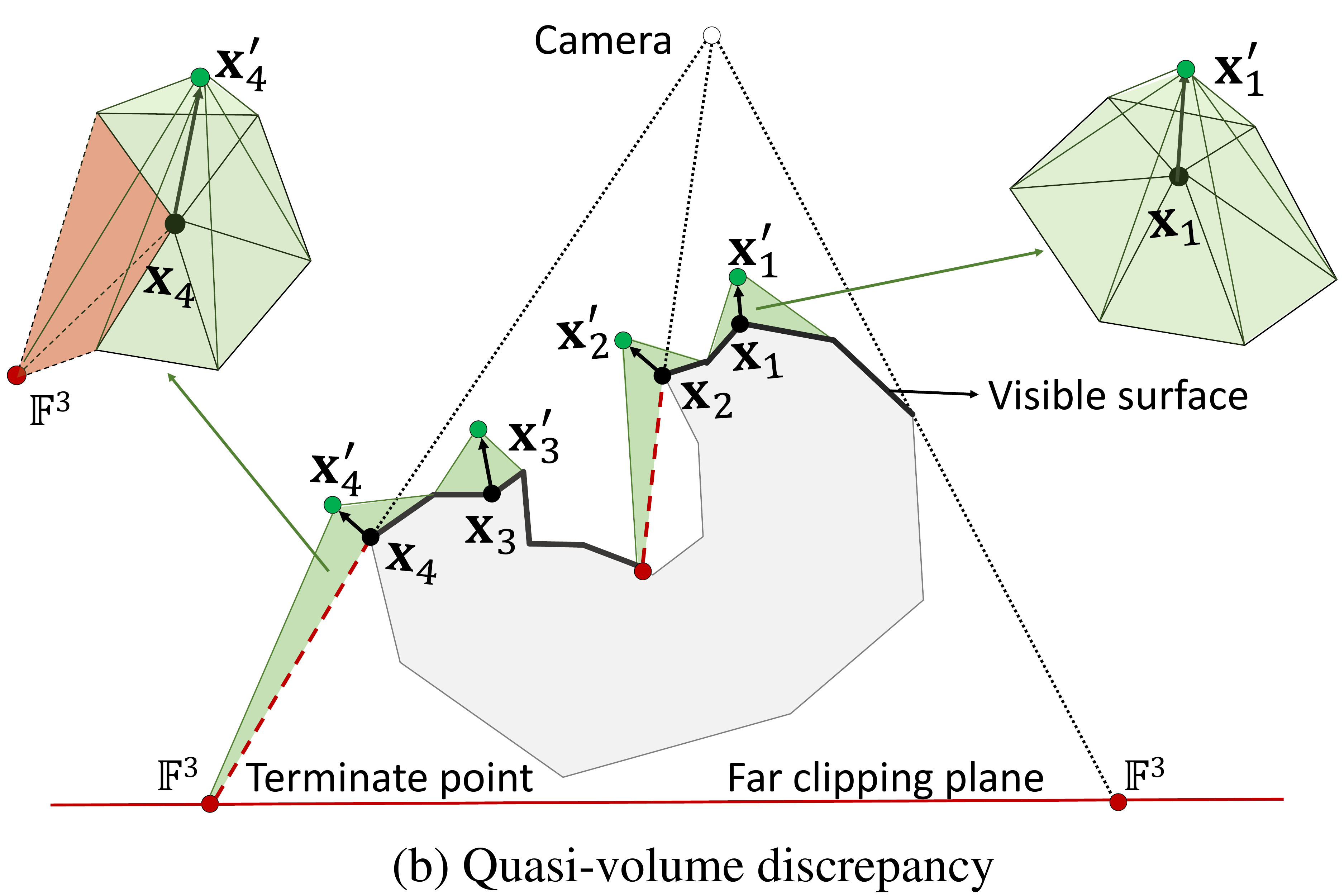} &
	\includegraphics[height=0.2\linewidth, trim={0cm 0cm 0cm 0cm}, clip]{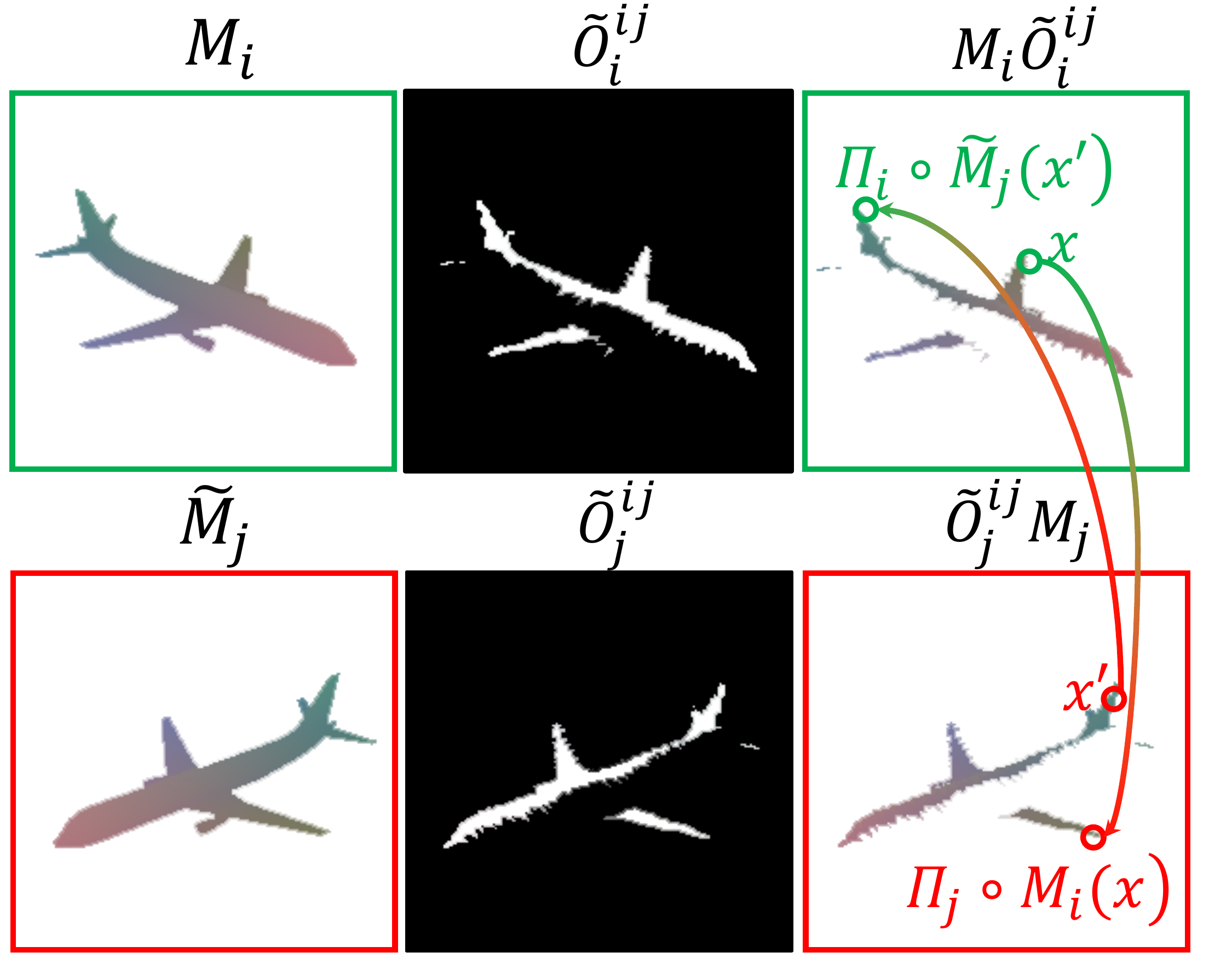}\\
	(a) Point-wise distance term. & (b) Quasi-volume term. & (c) Multi-view consistency term.
	\end{tabular}
	\caption{\textbf{Loss functions.} (a) Point-wise distances for 1-VPC. Induced by the same viewpoint, groundtruth and predicted points ($\widetilde{\mathbf{x}}_k$ and $\mathbf{x}_k$) stored in $x_k$ indicate the same surface point. (b) Quasi-volume discrepancy for typical examples. The local volume discrepancies of points around depth discontinuities, \eg, $\mathbf{x}_2$ and $\mathbf{x}_4$, are largely penalized by keeping fake connectivities (red dashed lines). (c) $\widetilde{O}_i^{ij}$ is the projection on view $i$ of the overlap region visible from view $i$ and $j$. $\widetilde{O}_j^{ij}$ is the projected overlap region on view $j$. Considering the overlap region, the multi-view consistency term minimizes the sum of distances between 3D points stored in pixels $x \in M_i$ and its reprojected pixel $\Pi_j \circ M_i(x) \in \widetilde{M}_j$, and verse visa.
	}
\label{fig:loss}
\end{figure*}

The points in groundtruth and predicted 1-VPC have a one-to-one mapping according to the definition of MVPC, illustrated in \fig{\ref{fig:loss}} (a). 2D pixels with equal 2D coordinates are defined to store the same surface point induced by the same viewpoint. 
Therefore, the sum of point-wise distances for groundtruth and predicted 1-VPC is the L2 loss. The total sum of point-wise distances of MVPC is given by:

\begin{eqnarray}
\label{eq:pt}
\mathcal{L}_{ptd} 
= \sum_{i}^N  \sum_{x \in M_i} ||M_i(x) - \widetilde{M}_i(x)||_2
\end{eqnarray}
where $x$ is a 2D pixel, $M_i(x)$ and $\widetilde{M}_i(x)$ denotes the 3D coordinates stored in predicted 1-VPC $M_i$ and groundtruth 1-VPC $\widetilde{M}_i$ at $x$. Here we take the visibility into account by setting $\widetilde{M}_i(x)$ to an infinite point $\mathbb{F}^3$ (a point at the far clipping plane practically). 

Neural networks tend to predict a mean shape averaging out the space of uncertainty using L2 or L1 loss. 
Point-wise distances neglecting local interactions do not fully express geometric discrepancy between surfaces.
Moreover, this metric may give rise to erroneous reconstructions around occluding contours, because minor errors on 2D projective planes lead to large 3D deviations at depth continuities.

\subsubsection{Quasi-volume term.}
\label{sec:volume} 
The limitation of point-wise distance metrics motivates us to formulate discrepancies over surfaces. Inspired by the volume-preserving constraints used in variational surface deformation~\cite{eckstein2007surfaceflow}, we propose a quasi-volume discrepancy metric to better describe the surface discrepancy. This term is able to characterize fine details and deal with occluding contours.

Let us first define the volume discrepancy between predicted and groundtruth continuous surfaces:
\begin{eqnarray}
\label{eq:volume_define}
\mathcal{L}_{vol}(\mathcal{S}, \widetilde{\mathcal{S}}) = \int _{\mathcal{S}} (\mathbf{x} - \widetilde{\mathbf{x}} ) \cdot \mathbf{n} \mathrm{d} {\mathbf{x}}
\end{eqnarray}
where $\mathbf{x}$ and $\widetilde{\mathbf{x}}$ are 3D points on predicted surface $\mathcal{S}$ and groundtruth surface $\widetilde{\mathcal{S}}$,
$\mathrm{d}\mathbf{x}$ is an area element of a surface and $\mathbf{n}$ is the outward normal to the surface at point $\mathbf{x}$, $(\cdot)$ denotes the inner product operator.

The discrete volume discrepancy of the MVPC representation can be deduced as:
\begin{eqnarray}
\label{eq:volume_vis}
\mathcal{L}_{vol} = \sum_{i}^N \sum_{x \in M_i}  \widetilde{V}_i(x) ( M_i(x)  - \widetilde{M}_i(x)) \cdot \widetilde{\mathbf{N}}_i(x) 
\end{eqnarray}
where $\widetilde{V}_i$ is the groundtruth visibility map and $\widetilde{\mathbf{N}}_i$ is the groundtruth area-weighted normal map of view $i$. 
Formally, for each pixel $x$ with its backprojected surface point $\mathbf{x}$, the normal map of view $i$ is $\mathbf{N}_i(x) = \sum_{\Delta \in \Omega(\mathbf{x})}|\Delta |\mathbf{n}(\mathbf{x})$, where $\Delta $ denotes a mesh triangle, $\Omega(\mathbf{x})$ contains the 1-ring triangles around point $\mathbf{x}$, $\mathbf{n}(\mathbf{x})$ is the outward normal at point $\mathbf{x}$.
A detailed proof is presented in the supplemental material.
Note that we use the groundtruth visibility $\widetilde{V}_i(\cdot)$, and thus add a cross entropy loss accounting for the visibilities.

\eqn{\ref{eq:volume_vis}} formulates the volume discrepancy for visible parts. We complement it with invisible parts, and name it as \emph{quasi-volume} discrepancy, as illustrated in \fig{\ref{fig:loss}} (b). We assign the background pixels with a terminate point $\mathbb{F}^3$ at far clipping plane. Thus, the points at boundaries of ${M}_i$ will achieve a large discrepancy gain, which means they are of high weights in the loss function, \eg, $\mathbf{x}_4$ in \fig{\ref{fig:loss}} (b). Similarly, points at occluding contours experience large volume loss, \eg, $\mathbf{x}_2$ in \fig{\ref{fig:loss}} (b). The quasi-volume term implicitly handles the challenges introduced by occluding contours.

\subsubsection{Multi-view consistency term.}
\label{sec:mv_consistency}

Partial surfaces of an object visible from different viewpoints may have overlap, which can be reached by letting points from different views attract one another.
The consistency serves as links between groundtruth and predicted 3D points stored in a pair of corresponding pixels from two different views. \fig{\ref{fig:loss}} (c) shows an example for two views. Note that the consistency only exists at overlap regions. We first compute the projected overlap region $\widetilde{O}_i^{ij}$ on view $i$ by rendering groundtruth 1-VPC $\widetilde{M}_j$ on view $i$, and get $\widetilde{O}_j^{ij}$ by rendering $\widetilde{M}_i$ on view $j$.
We minimize the sum of two distances between the stored 3D points in two corresponding pixels and their reprojected pixels in the other view.
For each pixel $x$ in $\widetilde{O}_i^{ij}$, the predicted 3D coordinate is $M_i(x)$. The reprojected pixel on view $j$ is $\Pi_j \circ M_i(x)$, where $\Pi_j$ denotes the projection matrix of view $j$.
Similarly, the pixel $x'$ at groundtruth 1-VPC $\widetilde{M}_j$ corresponds to the pixel $\Pi_i \circ \widetilde{M}_j(x')$ in the predicted 1-VPC $M_i$.
Therefore, the multi-view consistency term takes the form:
\begin{eqnarray}
\mathcal{L}_{mv} = \sum_{i,j} (\sum_{x \in \widetilde{O}_i^{ij}}||M_i(x) - \widetilde{M}_j(\Pi_j \circ   M_i(x))||_2  \nonumber \\
+\sum_{x \in \widetilde{O}_j^{ij}}||\widetilde{M}_j(x) - M_i(\Pi_i \circ \widetilde{M}_j(x))||_2)
\end{eqnarray}
The multi-view consistency term does not directly minimize distances between two predicted 1-VPCs but leverages the correspondences between predictions and groundtruths. This is because erroneous 3D coordinates in predictions will introduce false correspondences, resulting in divergence or falling into a trivial solution.
\section{Experiment}

\begin{table*}[t]
	\centering
	\caption{Quantitative comparison to the state-of-the-arts with per-category voxel IoU.}
	\label{tab:iou_multi}
	\scalebox{0.71}{
		\begin{tabular}{c|c||c|c|c|c|c|c|c|c|c|c|c|c|c|c}
			&	& plane & bench & cabinet & car   & chair & display & lamp  & speaker & firearm & couch & table & phone & vessel & mean  \\
			\hline
			\hline

			\multirow{4}*{\rotatebox[]{90}{\normalsize{voxel}}}	&R2N2\cite{choy20163dr2n2}(1 view)& 0.513 & 0.421 & 0.716   & 0.798 & 0.466 & 0.468   & 0.381 & 0.662   & 0.544   & 0.628 & 0.513 & 0.661 & 0.513  & 0.56  \\
			~&R2N2\cite{choy20163dr2n2}(5 views)& 0.561 & 0.527 & \bf{0.772}& \bf{0.836} & 0.550 & 0.565 & 0.421 & 0.717 &0.600 & 0.706 & 0.580 & 0.754 & 0.610 &  0.630 \\
			~&PTN-Comb\cite{yan2016ptn}      & 0.584 & 0.508 & 0.711 & 0.738 & 0.470 & 0.547 & 0.422 & 0.587 & 0.610 & 0.653 & 0.515 & 0.773 & 0.551 & 0.590  \\
			~&CNN-Vol\cite{yan2016ptn} & 0.575 & 0.514 & 0.697 & 0.735 & 0.445 & 0.539 & 0.386 & 0.548 & 0.603 & 0.647 & 0.514 & 0.769 & 0.5445 & 0.578  \\
			\hline
			\multirow{2}*{\rotatebox[]{90}{point}}		&Soltani\cite{soltani2017synthesizing} & 0.587 & 0.524 & 0.698   & 0.743 & 0.529 & 0.679   & 0.480  & 0.586   & 0.635   & 0.59  & 0.593 & 0.789 & 0.604  & 0.618 \\
			
			&Su\cite{fan2017point}  & 0.601 & 0.55  & 0.771   & 0.831 & 0.544 & 0.552   & 0.462 & \bf{0.737}   & 0.604   & \bf{0.708} & 0.606 & 0.749 & 0.611  & 0.640  \\
			\hline
			\multirow{3}*{\rotatebox[]{90}{Depth}} 
			&GeoLoss($N$=4)&0.655 &	0.578&	0.664&	0.709&	0.546&	0.653&	0.486&	0.573&	0.676&	0.630&	0.561&	0.783&	0.633&	0.627\\
			~&GeoLoss($N$=6)&0.624&	0.579&	0.677&	0.719&	0.543&	0.636&	0.498&	0.578&	0.682&	0.636&	0.548&	0.800&	0.643&	0.628\\
			~&GeoLoss($N$=8)&0.622&	0.576&	0.691&	0.724&	0.540&	0.643&	0.501&	0.590&	0.684&	0.647&	0.534&	0.788&	0.640&	0.629\\
			\hline
			\multirow{4}*{\rotatebox[]{90}{MVPNet}}		&PtLoss($N$=6) & 0.474 & 0.459 & 0.573   & 0.704 & 0.436 & 0.558     & 0.375 & 0.496   & 0.519   & 0.567 & 0.432 & 0.691 & 0.558  & 0.526 \\
			\cline{2-16}
			~&GeoLoss($N$=4)& 0.666 & 0.622 & 0.693 & 0.786 & 0.616 & 0.653 & 0.510 & 0.599 & \bf{0.696} & 0.690 & 0.635 & 0.811 & \bf{0.663} & 0.665\\
			
			~&GeoLoss($N$=6)& \bf{0.678} & \bf{0.623} & 0.685 & 0.788 & \bf{0.627} & \bf{0.681} &\bf{0.523} & 0.602 &0.693 & 0.701 & \bf{0.652} & \bf{0.814} & 0.659 & \bf{0.671}\\
			
			~&GeoLoss($N$=8)& 0.667 & 0.610  & 0.686   & 0.782 & 0.609 & 0.667   & 0.507 & 0.596   & 0.688   & 0.686 & 0.641 & 0.809 & 0.661  & 0.662
			
		\end{tabular}
	}
\end{table*}

\begin{table*}[]
	\centering
	\caption{Quantitative comparison to point-based methods using the chamfer distance metric. All numbers are scaled by 0.01.}
	\label{tab:distance}
	\scalebox{0.76}{
		\begin{tabular}{c||c|c|c|c|c|c|c|c|c|c|c|c|c|c}
			& plane & bench & cabinet & car   & chair & display & lamp  & speaker & firearm & couch  & table & phone & vessel & mean\\
			\hline
			\hline
			Su\cite{fan2017point} & 1.395      & 1.899      & 2.454      & 1.927      & 2.121      & 2.127      & 2.280      & 3.000       & 1.337      & 2.688      & 2.052      & 1.753      & 2.064      & 2.084  \\
			Lin\cite{lin2017learning}& 1.418      & 1.622      & 1.443      & 1.254      & 1.964      & 1.640      & 3.547      & 2.039       & 1.400      & 1.670      & 1.655      & 1.569      & 1.682      & 1.761  \\
			Soltani\cite{soltani2017synthesizing}&  0.167     &  0.165     & 0.122      & \bf{0.026} & 0.277   & 0.085  &  1.814   & 0.163   &  0.107  & 0.138   & 0.226  &  0.258     & 0.102      & 0.28    \\
			\hline
			MVPNet($N$=4) &  0.045   & 0.084 & 0.063 & 0.042 & 0.086 & 0.065 &　0.561 & 0.163 & 0.104 & 0.082 & 0.070 & 0.046 & 0.060 & 0.113 \\
			MVPNet($N$=6) & \bf{0.041} & \bf{0.079} & 0.060 & 0.041 & \bf{0.085} & 0.053 & \bf{0.421} & \bf{0.152} & \bf{0.093} & \bf{0.070} & \bf{0.069} & \bf{0.038} & \bf{0.050} & \bf{0.096}\\
			
			MVPNet($N$=8)   & 0.044 & 0.085 & \bf{0.058}   & 0.040  & 0.103 & \bf{0.050} & 0.494 & 0.153  & 0.113 & 0.083 & 0.075 & 0.039 & 0.059  & 0.107 
			
			
		\end{tabular}
	}
\end{table*}
\subsection{Implementation}
\label{sec:implementation}
We show the architecture of MVPNet in \fig{\ref{fig:net}}.
The input RGB image is of size $128 \times 128$. The output surface coordinate maps is of shape $N \times 128 \times 128 \times 4$. 
The encoder consists of five convolution (conv) layers with numbers of channels \{32, 64, 128, 256, 512\}, kernel sizes \{3, 3, 3, 3, 3\}, and strides \{2, 2, 2, 2, 2\}, and two fully connected (fc) layers with numbers of neurons \{4096, 2048\}. The camera matrix $\mathbf{c}$ is encoded with two fc layers with numbers of neurons \{64, 512\}. 
The decoder part takes the concatenated feature $(z,c_i)$ as input and generates a surface coordinate map for each viewpoint.
The structure of the decoder is mirrored to the encoder, consisting of two fc layers and five transposed-convolution (also known as ``deconv'') layers for up-sampling. 
We add the last conv layer with the number of channels 4 and kernel size 1 to generate 4-channel output.
Batch normalization~\cite{ioffe2015batch} is not performed because we observe the training process is smooth. Leaky ReLU activation with a negative slope of $0.2$ is applied after all conv layers except the last one which is followed by the tanh layer. 

We train the network with Tensorflow~\cite{abadi2016tensorflow} on a Nvidia TitanX GPU with a minibatch of 32. We use Adam optimizer \cite{kingma2014adam} with a learning rate of 0.0001. 
The training procedure takes 100,000 iterations. The coefficients $\alpha$ and $\beta$ of GeoLoss is set to 100 and 1 respectively after 10000 iterations and both to 0 before, because the initial point clouds are noisy and the computed volume discrepancy and consistency term are not reliable.

\subsection{Dataset}
We leverage the ShapeNet \cite{chang2015shapenet} dataset, which contains a large volume of clean CAD models for our experiments. We setup two datasets for single-class and multi-class cases. The chair category (\textit{ShapeNet-Chair}) is used for single-class processing since it is ubiquitously evaluated in previous methods. For the multi-class dataset, we use 13 major classes as the 3D-R2N2 \cite{choy20163dr2n2} set, listed in \tab{\ref{tab:iou_multi}}, named as \textit{ShapeNet-13}. 
The datasets are split into training and testing sets with the fraction $0.8/0.2$.

To obtain input RGB images, we render each 3D model for 24 viewpoints which are randomly sampled with an elevation ranging from (-20, 20), an azimuth ranging from (0, 360) degrees, and a radius ranging from (0.6, 2.3). Note that all models are normalized by their bounding spheres' radius.
\subsubsection{Viewpoint arrangement.}
For the viewpoint arrangement of the output MVPC, we approximately maximize the coverage of the ``mean'' shape (the unit sphere) of all objects with respect to $N$. The $N$ (4, 6, 8) viewpoints are located at vertices of a tetrahedron, octahedron, and cube, respectively. 
All the viewpoints look at the origin. Orthogonal projection is used to avoid additional perspective distortion. We calculate the average surface coverage by counting the number of visible points in groundtruth models, which are 97.2\%, 97.7\% and 98.0\% for $N$=4, 6, 8 respectively. The performances of different viewpoint settings are reported.

\subsection{Reconstruction Result}
Both qualitative and quantitative results of the reconstruction are presented. We compare our method to two collections of state-of-the-art methods according to the final result representations, namely, point clouds and volumetric grids.
\begin{figure*}
	\centering
	\includegraphics[width=1\linewidth]{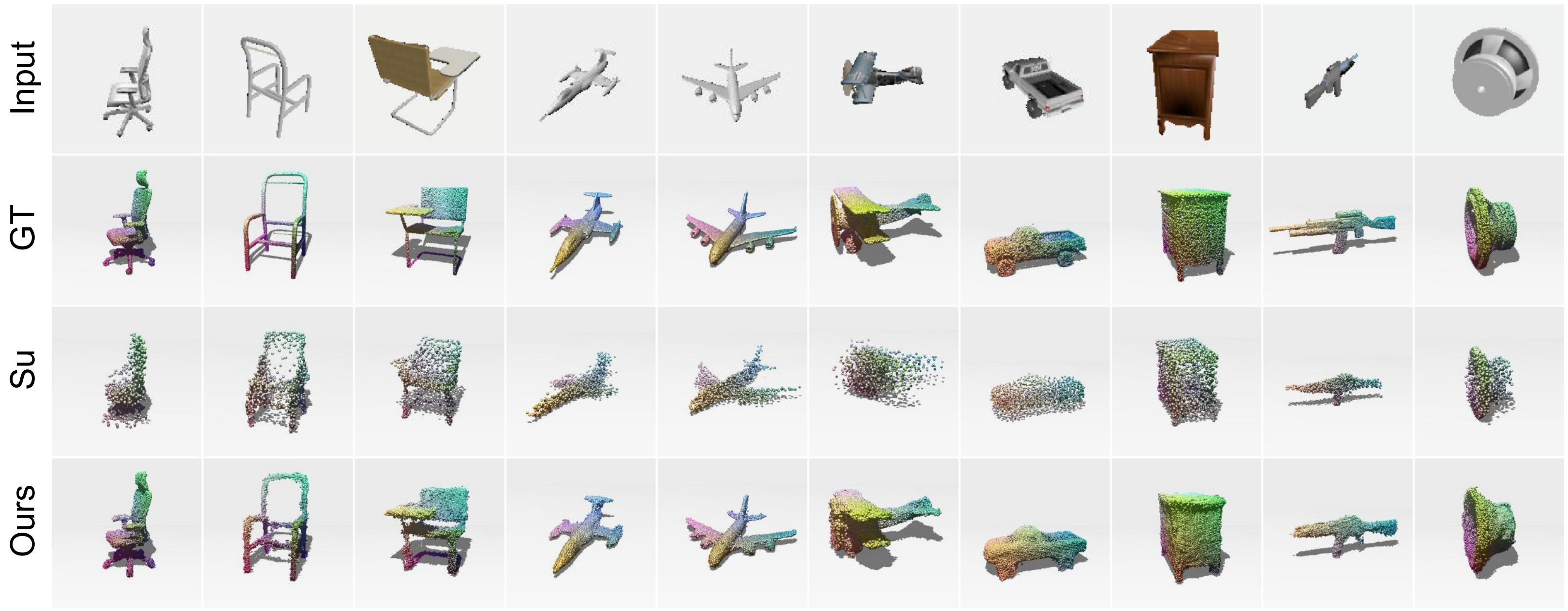}
	\caption{Qualitative comparison to point generation method. Compared with Su \etal \cite{fan2017point}, our method preserves more fine details and recovers better concave structures.}
	\label{fig:comparison}
\end{figure*}
\begin{figure*}[t]
	\centering
	\includegraphics[width=1\linewidth]{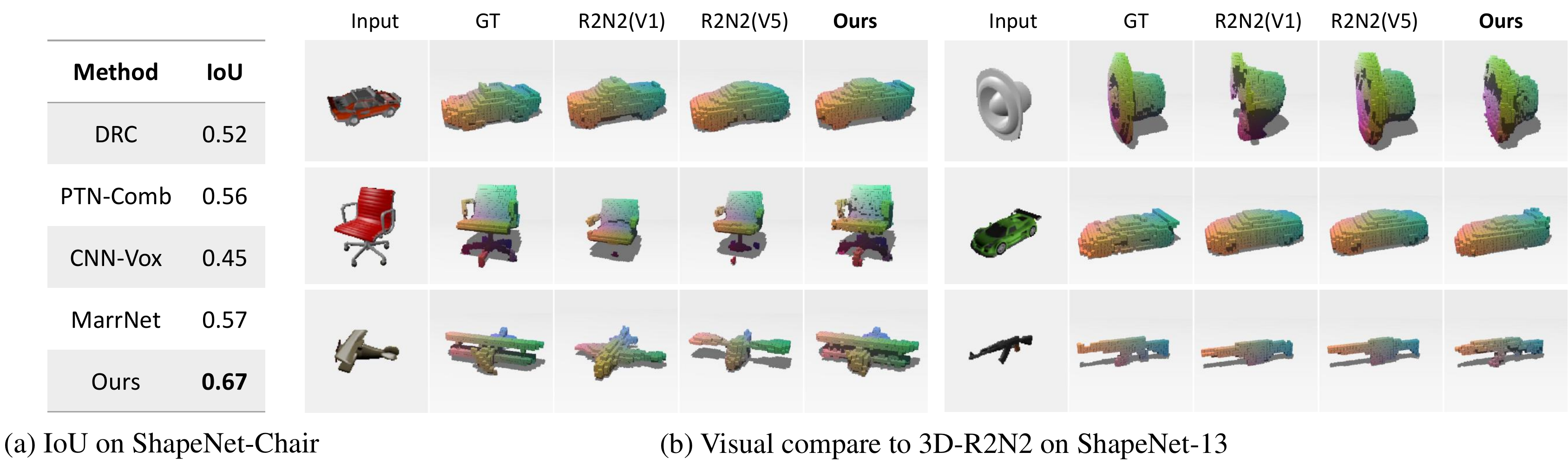}
	\caption{Comparison to voxel-based methods. (a) Quantitative comparison on ShapeNet-Chair. (b) Qualitative comparison to 3D-R2N2~\cite{choy20163dr2n2} on ShapeNet-13. Our method with a single image generates more detailed results than 3D-R2N2 with 5 input images (V5).
	}
	\label{fig:voxel_compare}
\end{figure*}
\subsubsection{Comparison to point generation methods.}

We compare our method to the state-of-the-art point generation methods using both an unordered point cloud representation~\cite{fan2017point} and view-based representations~\cite{soltani2017synthesizing,lin2017learning} on the ShapeNet-13 dataset.
Note that the methods proposed by Su \etal~\cite{fan2017point}, Lin \etal~\cite{lin2017learning} and us take a single RGB image as the input, while Soltani \etal~\cite{soltani2017synthesizing} use depth maps as input which may contain more geometric information.
We use Intersection-of-Union (IoU) of voxel occupancy for evaluating the reconstruction accuracy as most methods do.
The per-class IoU statistics are reported in \tab{\ref{tab:iou_multi}}. Our results of different numbers of viewpoints ($N$=4,6,8) are reported for discussion.
For a fair comparison, we adopt the model proposed in \cite{soltani2017synthesizing} without class supervision and post-processing.
All the results are generated from the trained model provided by the authors~\cite{soltani2017synthesizing,fan2017point}.

Our results using GeoLoss outperform the previous methods on 9 out of 13 classes with 6 and 8 viewpoints. The results of 6 viewpoints achieve best on 7 classes, which demonstrates that 6 viewpoints are sufficient to cover most objects and suppress error propagation in multi-view learning.
Note that our network with the GeoLoss achieves significantly better results on classes with a lot of thin and complex structures, such as planes (+17\%), chairs(+7\%), and lamps(+6\%). This is because our method exacts at capturing fine details by minimizing the GeoLoss which interprets the variance over 3D surfaces rather than sparse points or 2D projective planes. The importance of GeoLoss is demonstrated by comparing with the results using only point-wise distance term (PtLoss). We find that the results from GeoLoss are about 10\% higher than the ones from PtLoss with 6 viewpoints (best in GeoLoss).
We use GeoLoss in the following experiments.
Our reconstruction accuracy for classes with simple structures, \eg, cabinet, car and display, are slightly lower than Su~\cite{fan2017point} and Soltani~\cite{soltani2017synthesizing}, because their net architectures are more complex (``hourglass'' structure in Su~\cite{fan2017point} and ``ResNet blocks'' in Soltani~\cite{soltani2017synthesizing}) and predict better ``mean'' shapes.

To evaluate the faithfulness of the generated point to the groundtruth surface, we compute the Chamfer Distance (CD)~\cite{fan2017point} between the prediction and the densely sampled points from groundtruth meshes. CD is a common measure of the distance between two point sets, which is defined by summing up the distances between each source point to its nearest point in the target point set. The groundtruth points of size 100,000 are uniformly sampled on the surface. The CD evaluation is reported in \tab{\ref{tab:distance}}. Our method is superior to the previous methods on most classes (12/13) by a large margin. Same as in IoU evaluation, 6 viewpoints get the best on 10 classes. The multi-view point clouds generated by our network possess high density and the geometric loss enforces local spatial coherence. The unordered point generation method~\cite{fan2017point} gets sparse point clouds which are limited to characterizing enough details, leading to large chamfer distances. The method proposed by Soltani \etal \cite{soltani2017synthesizing} also obtains small distances since it generates points with many more (20) depth maps. 


For qualitative comparison, we present several typical examples in \fig{\ref{fig:comparison}}. Our method is able to produce much denser points ($\sim$15k), while the method proposed by Su \cite{fan2017point} limits the point cloud size to 1024. Our method is superior in recovering fine details (see chair backs, plane tails and car wheels) and dealing with concave structures, such as car trunk and two layers of plane wings. The geometric loss that handles occlusion encourages the improvement on concave shapes. More results of ours are shown in supplemental materials.

\subsubsection{Comparison to voxel-based methods.}
We compare the proposed method to the state-of-the-art voxel-based methods, \ie,
3D-R2N2~\cite{choy20163dr2n2}, DRC~\cite{tulsiani2017drc}, two models of PTN~\cite{yan2016ptn} (PTN-Comb, CNN-Vox), and MarreNet~\cite{wu2017marrnet}. 
These methods directly use 3D volumetric representation and usually compute the IoU for evaluation. 
Since our results form dense point clouds, we convert them to $(32\times32\times32)$ grids as Su~\etal~\cite{fan2017point} do. 
For single class model, our method achieves much higher IoU (0.667) than the highest IoU (0.57) among the state-of-the-art methods on the ShapeNet-Chair dataset, shown in \fig{\ref{fig:voxel_compare}} (a).
For the multi-class results, we report per-category IoU in \tab{\ref{tab:iou_multi}} on ShapeNet-13 dataset. The qualitative comparison to 3D-R2N2 is shown in \fig{\ref{fig:voxel_compare}} (b). We show that our method preserves more fine details, such as legs of chairs, wings of planes, and holders of firearms.

\subsubsection{Comparison to depth regression.}
Here we show our findings that directly regressing 3D coordinates has advantages over regressing depths.
To compare 3D coordinates and depth regression, we adopt the same network architecture but the last layer and use the same GeoLoss. The channel numbers of the last layers are 3 and 1 for regressing coordinates and depths respectively.
As reported in \tab{\ref{tab:iou_multi}}, the depth regression generates reasonable results, but the accuracy is about 4\% lower than coordinate regression.
This is because searching a gradient decent move in 3D space with an arbitrary direction is more flexible and stable than searching in one fixed direction considering the loss of the 3D space (rather than 1D depth loss), especially on occluding contours. 
With the same 3D volume loss descent, the 3D point needs a small move in the steepest direction, while the depth needs an extremely large move in the fixed direction.
Thus, 3D coordinates are easier to learn than depths.
In addition, the complexity does not grow much because only the last layer of the decoder is different. 
\begin{figure}
	\centering
	\includegraphics[width=1\linewidth]{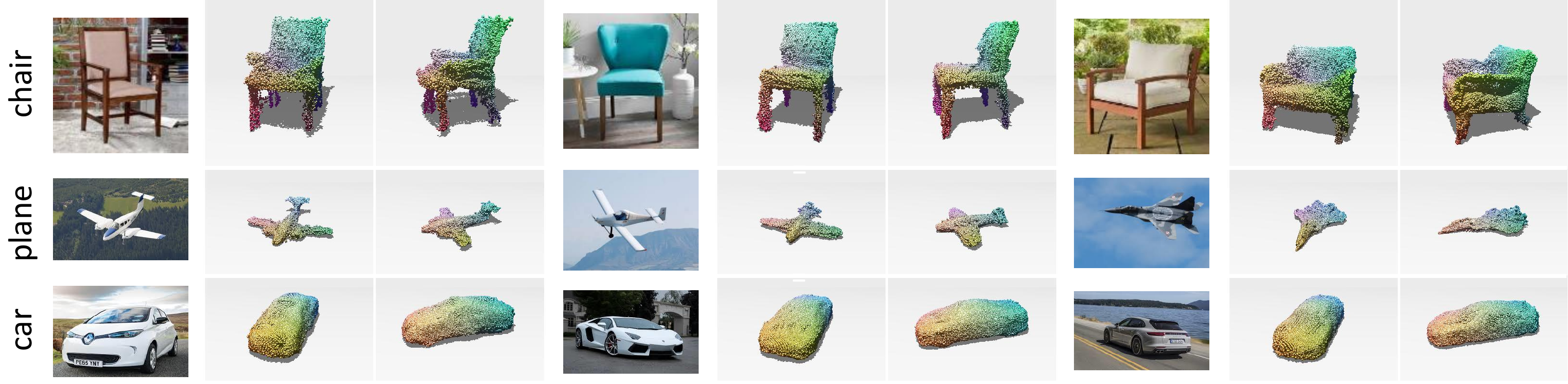}
	\caption{Reconstruction results on real word data.}
	\label{fig:realim}
\end{figure}
\begin{figure}
	\centering
	\includegraphics[width=1\linewidth]{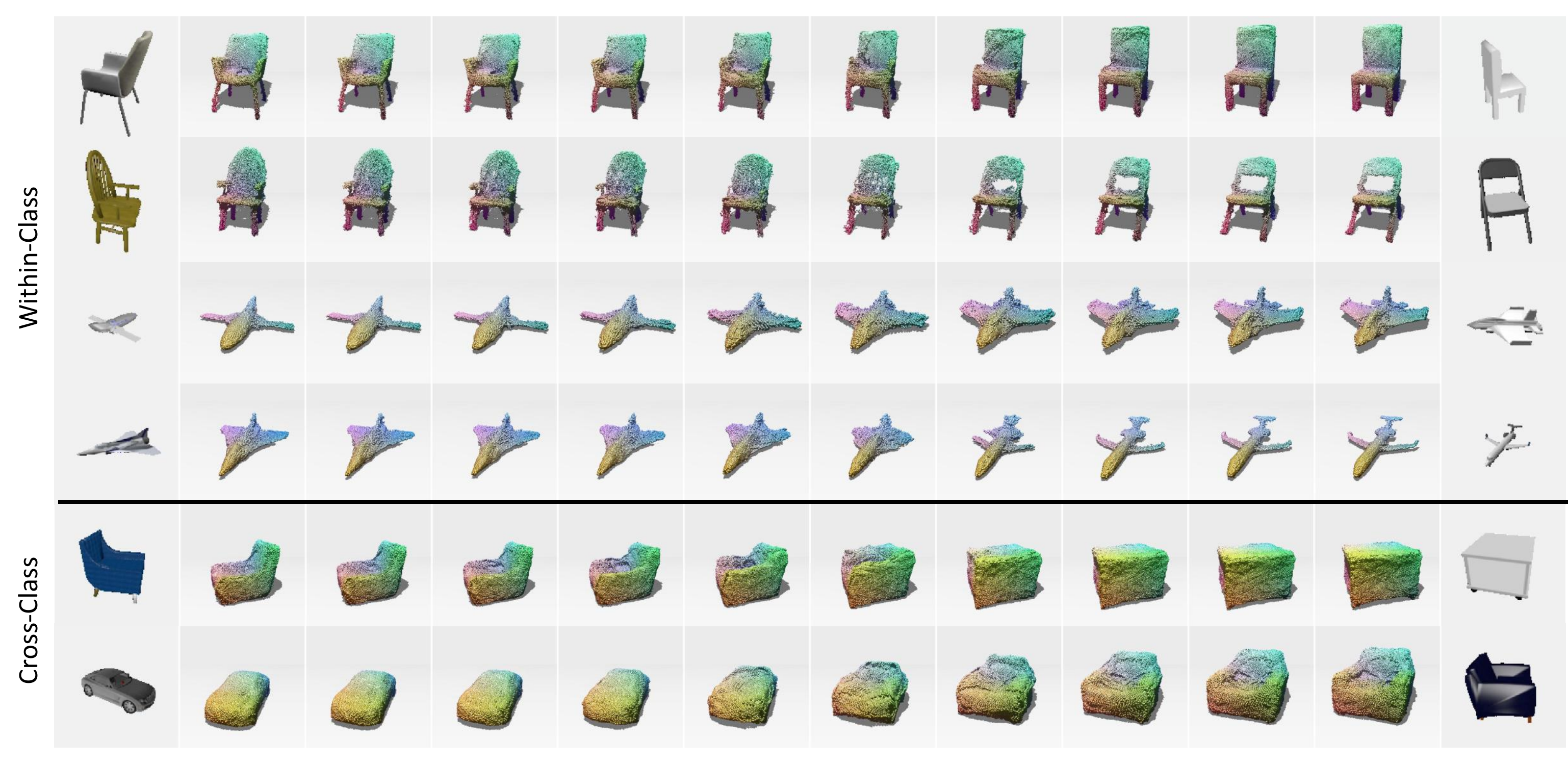}
	\caption{Reconstructions for linear interpolation of two learned latent features within and across classes respectively.}
	\label{fig:interpolation_classification}
\end{figure}

\subsubsection{Results on real dataset.}
We show our model works well on natural images without additional input.
To adjust our model to real-world images, we synthesize the training data by augmenting the input images with random crops from the PASCAL VOC 2012 dataset~\cite{everingham2011pascal} as \cite{tatarchenko2016mv3d} do. We show that the proposed method yields reasonable results in \fig{\ref{fig:realim}}.

\subsubsection{Application.}

We show the generative representation of the learned features using linear interpolation in \fig{\ref{fig:interpolation_classification}}. We can see clear and gradual transitions of the generated point clouds, indicating the learned feature space to be sufficiently representative and smooth. 
More results of discriminative representations are presented in the supplemental material.

\section{Conclusions}
We have presented the MVPNet for regressing dense 3D point clouds of an object from a single image. The point regression achieves state-of-the-art performance resorting to the MVPC representation and the geometric loss. 
The MVPC express an object's surface with view-dependent point clouds that are embedded in regular 2D grids, which easily fit into CNN-based architectures. Also, the one-to-one mapping from 2D pixels to reprojected 3D points makes these points in 1-VPC ordered, which accelerate the loss computation. 
Although the dimension of the data embedding space is reduced from 3D space to 2D projective planes, we propose the geometric loss that integrates variances over the 3D surfaces instead of the 2D projective planes. The experiments demonstrate the geometric loss significantly improves the reconstruction accuracy.

\par\vfill\par

\clearpage

\bibliographystyle{aaai}
\bibliography{reference}

\clearpage

\end{document}